\documentclass[10pt,twocolumn,letterpaper]{article}

\usepackage{cvpr}              

\usepackage{graphicx}
\usepackage{amsmath}
\usepackage{amssymb}
\usepackage{booktabs}

\usepackage{amsfonts}
\usepackage{capt-of}
\usepackage{etoolbox}
\usepackage{multirow}
\usepackage{tablefootnote}

\usepackage[pagebackref,breaklinks,colorlinks]{hyperref}

\usepackage[capitalize]{cleveref}
\crefname{section}{Sec.}{Secs.}
\Crefname{section}{Section}{Sections}
\Crefname{table}{Table}{Tables}
\crefname{table}{Tab.}{Tabs.}

\def\cvprPaperID{9} 
\def\confName{CVPR}
\def\confYear{2023}

\newcommand{\bhline}[1]{\noalign{\hrule height #1}}

\begin{document}
\title{Memory Efficient Diffusion Probabilistic Models via Patch-based Generation}

\author{
    \text{Shinei Arakawa}${}^1$\quad
    \text{Hideki Tsunashima}${}^1$\quad
    \text{Daichi Horita}${}^2$\quad
    \text{Keitaro Tanaka}${}^1$\quad
    \text{Shigeo Morishima}${}^3$\\
    ${}^1$Waseda University\quad
    ${}^2$The University of Tokyo\\
    ${}^3$Waseda Research Institute for Science and Engineering
}
\maketitle
    
    \begin{abstract}
\vspace{-1mm}
Diffusion probabilistic models have been successful in generating high-quality and diverse images.
However, traditional models, whose input and output are high-resolution images, suffer from excessive memory requirements, making them less practical for edge devices.
Previous approaches for generative adversarial networks proposed a patch-based method that uses positional encoding and global content information.
Nevertheless, designing a patch-based approach for diffusion probabilistic models is non-trivial.
In this paper, we present a diffusion probabilistic model that generates images on a patch-by-patch basis.
We propose two conditioning methods for a patch-based generation.
First, we propose position-wise conditioning using one-hot representation to ensure patches are in proper positions.
Second, we propose Global Content Conditioning (GCC) to ensure patches have coherent content when concatenated together.
We evaluate our model qualitatively and quantitatively on CelebA and LSUN bedroom datasets and demonstrate a moderate trade-off between maximum memory consumption and generated image quality.
Specifically, when an entire image is divided into $2\times2$ patches, our proposed approach can reduce the maximum memory consumption by half while maintaining comparable image quality.
\end{abstract}

    \section{Introduction}
    \label{sec:intro}
    \begin{figure}[!t]
    \begin{center}
        \centering
        \includegraphics[width=0.4\textwidth]{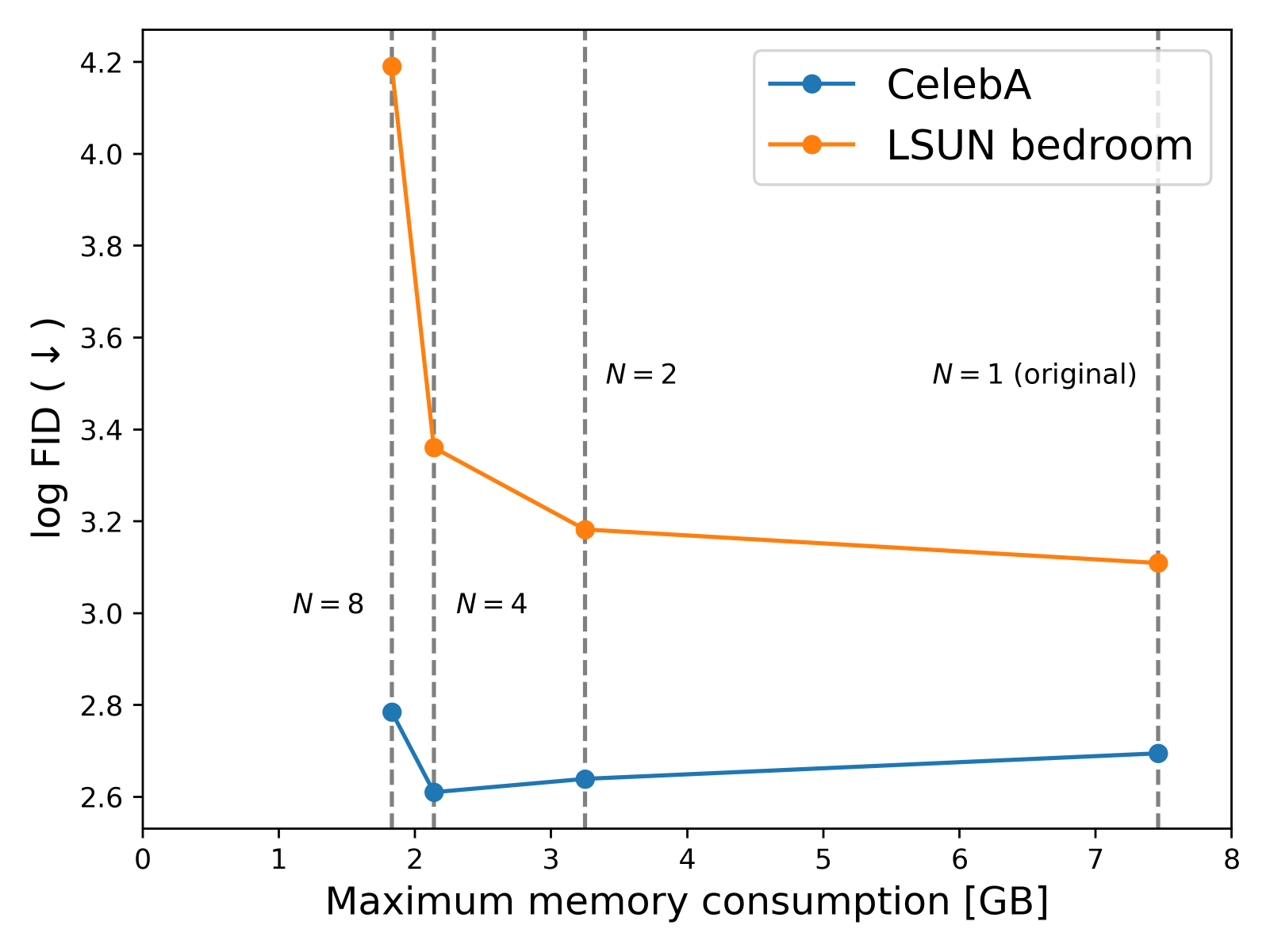}
        \vspace{-4mm}
        \caption{Trade-off between the maximum memory consumption and the generative quality measured by (log) FID.
        $N$ indicates the number of patch divisions for both vertical and horizontal directions. 
        Our proposed method demonstrates moderate trade-off.}
        \label{fig:FID_vs_memory}
    \end{center}
    \vspace{-8mm}
\end{figure}
Diffusion Probabilistic Models~(DPMs)~\cite{original_DDPM, DDPM, DDPM++} have achieved remarkable success in high-resolution image synthesis~\cite{ImprovedDDPM,ADM}, image editing~\cite{SDEdit, ILVR}, and super-resolution~\cite{SR3}.
Stable Diffusion (SD)~\cite{LDM} is a popular text-to-image diffusion model trained on a large-scale text-image paired dataset, LAION5B~\cite{LION5B}.
SD outperforms other generative models, including GANs~\cite{GAN, StyleGAN, StyleCLIP} and VAEs~\cite{VAE}, in both generative quality and diversity.
Despite being able to generate images on commercially available GPUs, SD faces challenges when generating high-resolution images of $512^2$ or executing on edge devices such as smartphones due to GPU memory limitations.

To address the similar limitation for GANs, COCO-GAN~\cite{COCO-GAN} proposed a memory-efficient approach by parallelly generating patches and concatenating them to generate a whole image.
COCO-GAN introduced a position embedding mechanism to specify patch locations and an information-sharing mechanism to ensure coherent image content across patches.
However, memory-efficient methods for DPMs have not been sufficiently discussed.

In this study, we present a DPM that generates images on a patch-by-patch basis to reduce memory consumption during inference.
Patch-based generation effectively reduces the size of the input to the self-attention mechanism~\cite{transformer}, which is commonly used in the network of DPMs and can be a bottleneck for memory consumption.
Consequently, patch-based generation is capable of significantly reducing the maximum amount of memory required for computation.

\begin{figure*}[t]
    \centering
    \includegraphics[width=0.70\textwidth]{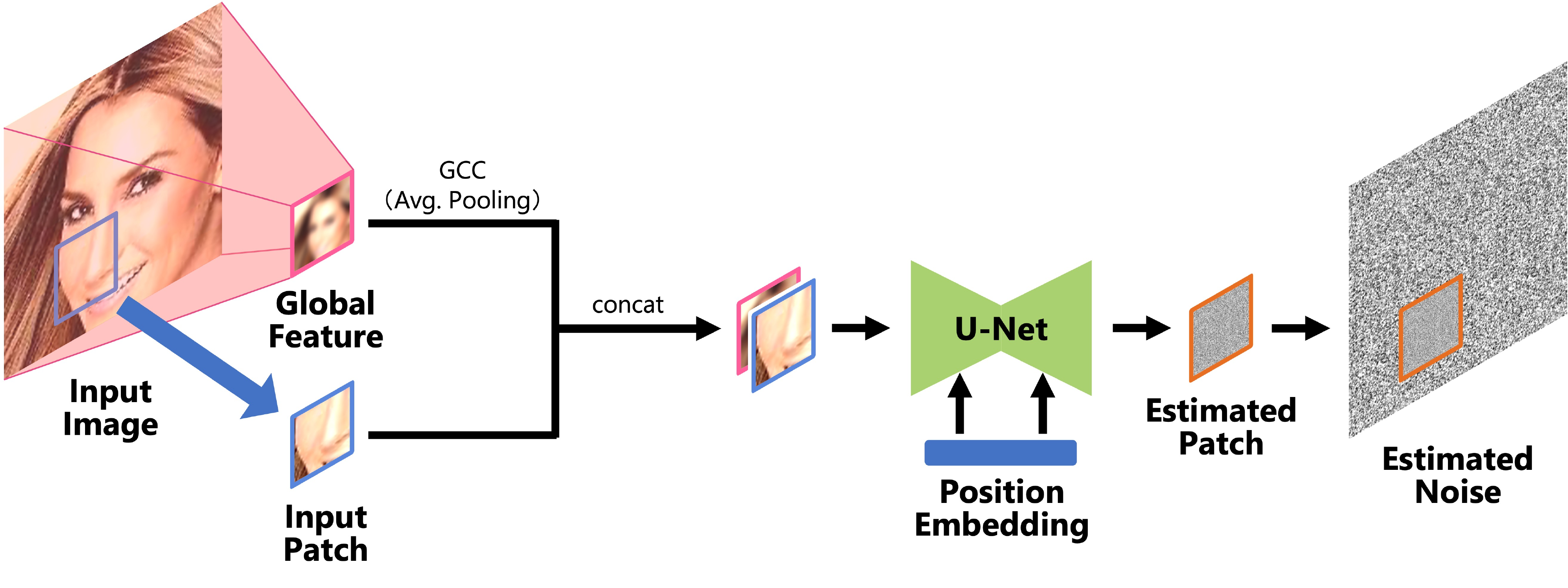}
    \vspace{-2mm}
    \caption{An overview of our proposed method. Global Content Conditioning~(GCC) extracts the global content feature from the entire image.
    After that, the global content feature is concatenated with the cropped patch and fed into the U-Net conditioned by patch position embedding.
    Our approach enables DPMs to generate images patch-by-patch, achieving high memory efficiency during inference. 
    }
    \vspace{-3mm}
    \label{fig:teaser}
\end{figure*}
Specifically, our study introduces two novel conditioning methods to enable DPMs to generate images patch-by-patch.
The first method conditions location information that specifies the position of the generated patch within the entire image.
We embed the position using a one-hot representation to enable proper conditioning.
The second method conditions the content information of the entire image to ensure coherence in the generated patches when rearranged.
To embed content information, we propose Global Content Conditioning (GCC), which first downsamples the entire image to get content information and then conditions the network by concatenating the extracted content feature with patches.
GCC can be implemented without any additional trainable parameters.
Through these conditioning methods, the proposed approach significantly reduces memory consumption during inference without dramatically increasing the number of network parameters.
An overview of our proposed method is illustrated in \cref{fig:teaser}.

We evaluated our proposed method on CelebA~\cite{CelebA} and LSUN bedroom~\cite{LSUN} datasets, which consist of face and bedroom images, respectively.
Our experimental results shed light on the trade-off relationship between generation quality and maximum memory consumption during inference, shown in \cref{fig:FID_vs_memory}.
Specifically, our approach can reduce the maximum memory consumption by half while maintaining comparable image quality when dividing an entire image into $2\times2$ patches. 

    \section{Background}
    \label{sec:background}
    Sohl-Dickstein et al.~\cite{original_DDPM} introduced Diffusion Probabilistic Models (DPMs) as a deep generative method. 
Ho et al.~\cite{DDPM} subsequently extended the model and achieved high generative performance on diverse datasets (e.g., CIFAR10~\cite{CIFAR10}, CelebA-HQ~\cite{CelebA-HQ}, and LSUN~\cite{LSUN}).
DPMs consist of forward and reverse processes. 
In this section, we provide a detailed description of each process.

\vspace{1mm}
\noindent{}\textbf{Forward Process}~
In DPMs, an image $\mathbf{x}_{0}$ sampled from data distribution $p(\mathbf{x})$ is perturbed iteratively by adding Gaussian noise. 
We define each diffusion step as $t\in\{1, \dots, T\}$, with the perturbed latent image denoted as $\mathbf{x}_t$. 
The gaussian transition is given by,
\begin{equation}
    \centering
    q(\mathbf{x}_{t}|\mathbf{x}_{t-1})=\mathcal{N}(\mathbf{x}_{t}|\sqrt{1-\beta_{t}}\mathbf{x}_{t-1},\beta_{t}\textbf{I}),
    \label{eq:forward_diffusion_kernel}
\end{equation}\noindent
where $\beta_{t}$ is the variance parameter at step $t$, determined such that $\mathbf{x}_{T}\approx\mathcal{N}(\mathbf{0},\textbf{I})$. 
Using \cref{eq:forward_diffusion_kernel}, we can write the conditional distribution of the latent variable $\mathbf{x}_t$, given the raw image $\mathbf{x}_0$, as a closed form,
\begin{equation}
    q(\mathbf{x}_{t}|\mathbf{x}_{0})=\mathcal{N}(\mathbf{x}_{t}|\sqrt{\bar{\alpha}_{t}}\mathbf{x}_{t-1},(1-\bar{\alpha}_{t})\textbf{I}),
    \label{eq:forward_diffusion_cumulative}
\end{equation}\noindent
where $\alpha_{t}=1-\beta_{t}$ and $\bar{\alpha}_{t}=\prod^{t}_{s=1}{\alpha_{s}}$. 
Given a real image $\mathbf{x}_0$, we can compute the perturbed image $\mathbf{x}_{t}$ at any diffusion step. 
This transition from data distribution to Gaussian distribution through repeated Gaussian noise addition constitutes the forward process.

\vspace{1mm}
\noindent{}\textbf{Reverse Process}~
The reverse process reproduces the data distribution by iterative denoising from a Gaussian distribution.
Starting from $p(\mathbf{x}_{T})=\mathcal{N}(\mathbf{0},\textbf{I})$, the iterative denoising is performed using the equation,
\begin{equation}
    \mathbf{x}_{t-1}
    =
    \frac{1}{\sqrt{\alpha_{t}}}
    \left(
        \mathbf{x}_{t}-
        \frac{\beta_{t}}{\sqrt{1-\bar{\alpha}_{t}}}\boldsymbol{\epsilon}_{\theta}(\mathbf{x}_{t},t)
    \right)+
    \sigma_{t}\mathbf{z},
    \label{eq:reverse_sampling_equation}
\end{equation}\noindent
where $\sigma_{t}=\sqrt{\beta_{t}}$ and $\mathbf{z}\sim\mathcal{N}(\mathbf{0},\textbf{I})$.
The noise added to the latent image $\mathbf{x}_t$ is approximated by a neural network denoted as $\boldsymbol{\epsilon}_{\theta}(\mathbf{x}_{t}, t)$.
U-Net~\cite{unet} is commmonly used for the neural network $\boldsymbol{\epsilon}_{\theta}(\mathbf{x}_{t}, t)$.

\vspace{1mm}
\noindent{}\textbf{Loss Function}~
The loss function for DPMs is expressed as a squared error,
\begin{equation}
    L=
    \mathbb{E}_{t,\mathbf{x}_{0},\boldsymbol{\epsilon}}
    \left[
        \left\|
            \boldsymbol{\epsilon}-
            \boldsymbol{\epsilon}_{\theta}(
                \sqrt{\bar{\alpha}_t}\mathbf{x}_{0}+
                \sqrt{1-\bar{\alpha}_t}\boldsymbol{\epsilon},
                t
                )
        \right\|^{2}
    \right].
    \label{eq:loss_function}
\end{equation}\noindent
In this formulation, the network of DPMs is trained by minimizing the squared error between the added noise in the forward process and the estimated noise.

    \section{Proposed Method}
    \label{sec:proposed_method}
    The network used in DPMs requires a significant amount of memory for computation since both input and output are high-dimensional images.
Our objective in this study is to reduce memory consumption during inference by partitioning images into patches and reducing the size of input and output variables passed to the network.
In this section, we describe a two-step approach for patch-by-patch generation.
First, we introduce a method for extracting patches from the entire image.
Next, we propose two conditioning methods that provide each location and content information required for generating patches.

\subsection{Patch Partitioning Method}
\begin{figure}[t]
    \begin{center}
        \centering
        \includegraphics[width=0.33\textwidth]{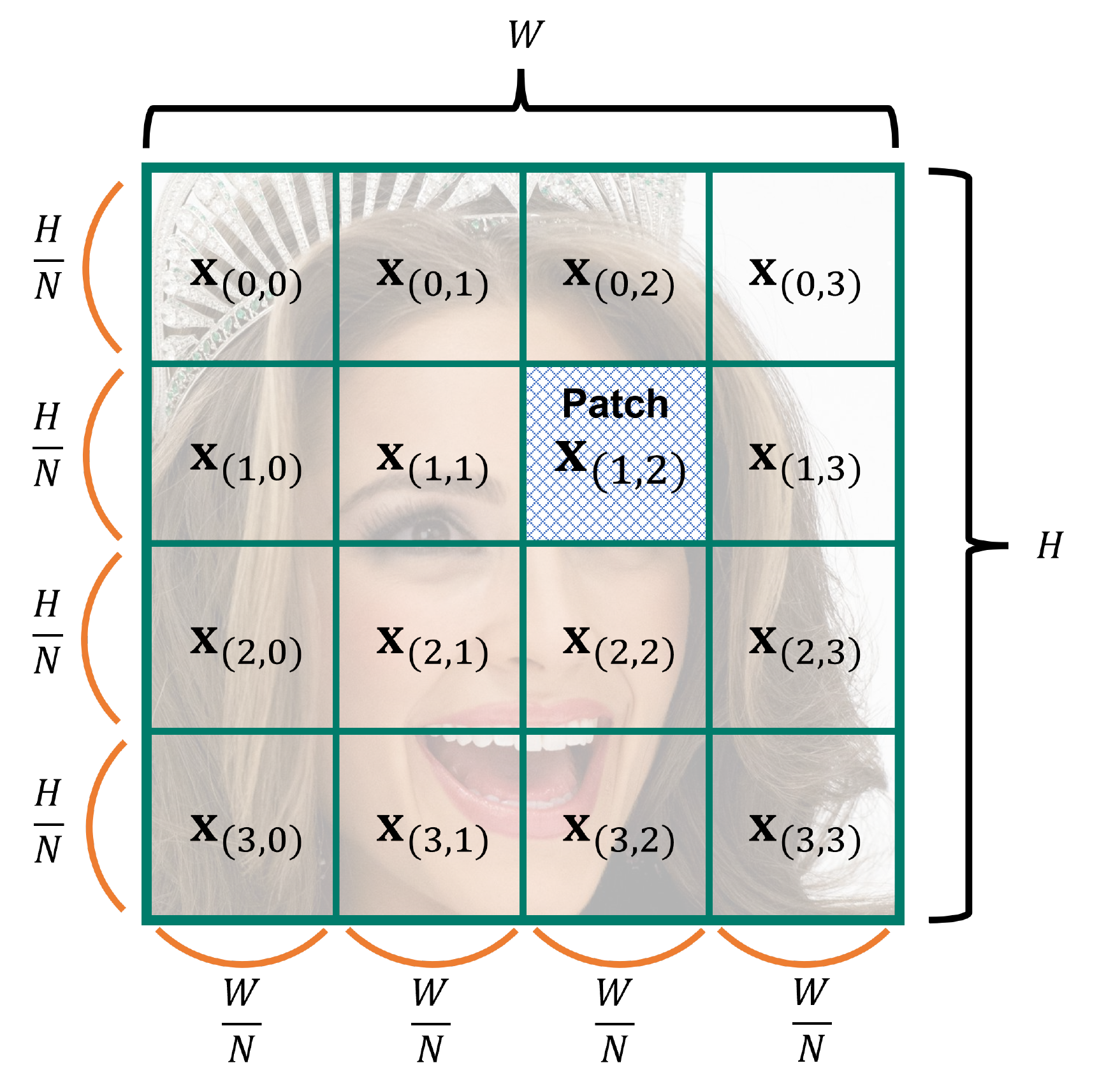}
        \vspace{-2mm}
        \caption{An example of the patch partitioning ($N=4$). The entire image is divided into $N^2=16$ patches, without any overlap.}
        \label{fig:patch_partitioning_method}
    \end{center}
    \vspace{-5mm}
\end{figure}
In conventional DPMs, the entire noisy image is passed into the network, and the amount of noise is estimated by a single forward propagation.
Our patch-by-patch generation method divides the input whole image into patches, passes the patches to the network individually, and rearranges the output patches to obtain the overall output.
Specifically, the input image $\mathbf{x}$ is divided into patches $\mathbf{x}_{(i,j)}$ in a grid-like manner with no overlap.
An example of the proposed patch partitioning method is shown in \cref{fig:patch_partitioning_method}.
We define the resolution of the input image $\mathbf{x}$ as $(H,W)$.
If the image is divided into $N$ equal parts in each direction, the resolution of a patch is expressed as $(H',W')=(H/N,W/N)$.
With the number of divisions $N$, the range of indices $i$ and $j$ is defined as $0 \leqq i,j \leqq N - 1$.

\subsection{Conditioning Location Information}
As the divided patches are independently input into the network, it is essential to provide patch location information to the network.
We address this by defining an index $s$ to specify the position of each patch uniquely.
We define the index $s$ for $N$-divided patches $\mathbf{x}_{(i,j)}$ as $s=i \times N + j$.
After converting the index $s$ to a one-hot representation (e.g., [0, 0, 1, 0]), it is passed to a fully connected layer, and the output is concatenated with the embedding features of the diffusion step (corresponding to the position embedding in \cref{fig:teaser}) and fed into the network.
This approach conditions the network with the relative position of an input patch.

\subsection{Conditioning Content Information}
We further propose Global Content Conditioning~(GCC) to address the lack of content information shared between independently processed patches in the network.
GCC extracts the overall content information $\mathbf{g}\in\mathbb{R}^{H'\times W'}$ through average pooling of the entire image $\mathbf{x}\in\mathbb{R}^{H\times W}$.
Then $\mathbf{g}$ is concatenated with the split patch $\mathbf{x}_{(i,j)}$ in the channel direction and input into the network.
GCC is parameter-free, and it can be implemented without adding model complexity.
This method is motivated by the technique proposed by Saharia et al.~\cite{SR3}, which demonstrates effective conditioning through simple concatenation in the channel direction for conditioning image features.

    \section{Experiments}
    \label{sec:experiments}
    \begin{figure}[t]
    \begin{center}
        \includegraphics[width=0.48\textwidth]{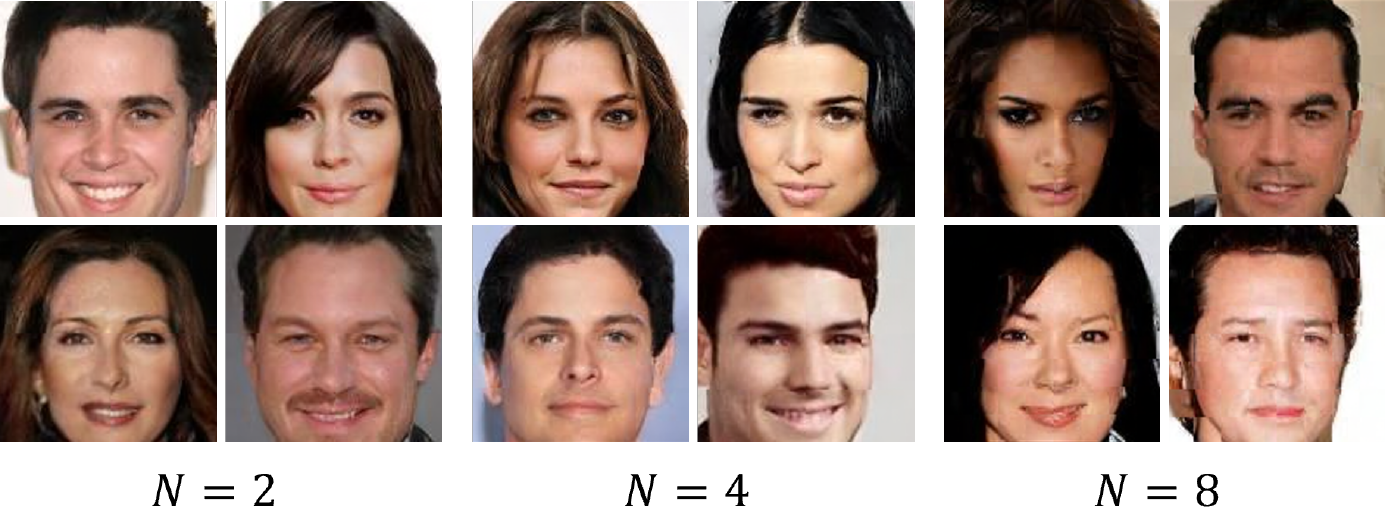}
        \vspace{-8mm}
        \caption{Generated images from our proposed method trained on CelebA dataset~\cite{CelebA}.}
        \label{fig:sampled_CelebA}
    \end{center}
    \vspace{-5mm}
\end{figure}
\begin{figure}[t]
    \begin{center}
        \includegraphics[width=0.48\textwidth]{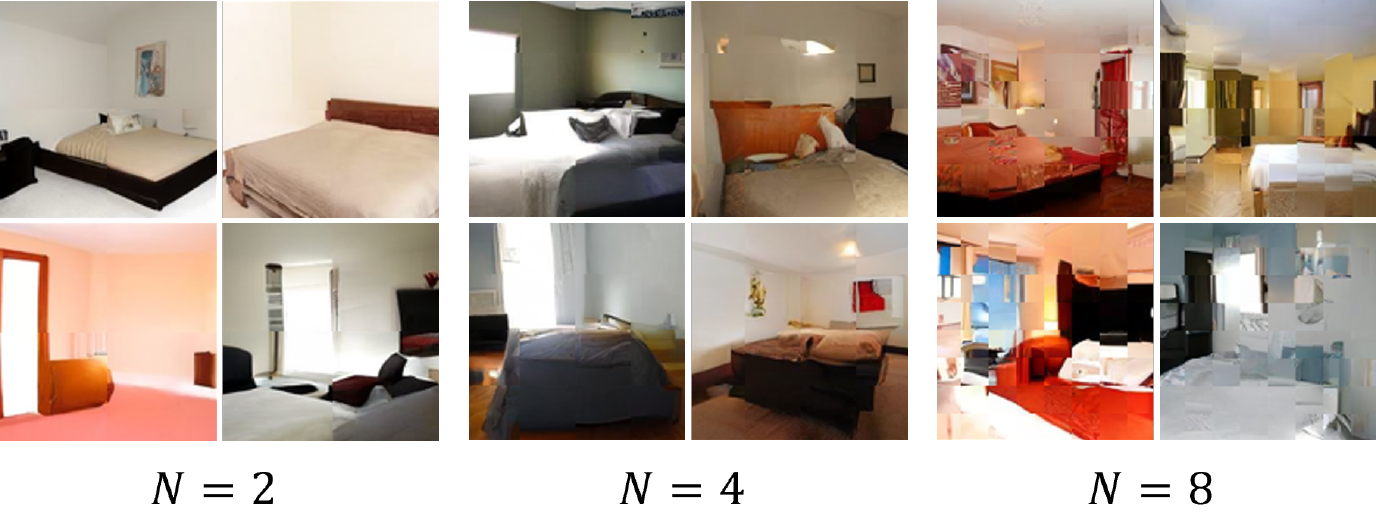}
        \vspace{-8mm}
        \caption{Generated images from our proposed method trained on LSUN bedroom dataset~\cite{LSUN}.}
        \label{fig:sampled_LSUN}
        \vspace{-8.0mm}
    \end{center}
\end{figure}

\subsection{Experiment Settings}
\vspace{1mm}\noindent{}\textbf{Dataset}~
We conducted experiments on two datasets: CelebA~\cite{CelebA}, consisting of approximately 200K face images, and LSUN bedroom~\cite{LSUN}, consisting of approximately three million bedroom images.
We preprocessed all images to be $128\times128$ pixels in size.
For CelebA, as the images were already square, we just downsampled them to $128^2$.
For LSUN bedroom, we first cropped the images with the short side resolution and then downsampled the cropped images to $128^2$.
The datasets were split into train, validation, and test sets, with proportions of $64\%:16\%:20\%$.

\vspace{1mm}\noindent{}\textbf{Training}~
The number of diffusion steps $T$ was fixed at 1000.
The mini-batch sizes were 64 for both CelebA and LSUN bedroom.
The learning rate was set to $1\times 10^{-4}$, and we chose Adam~\cite{Adam} as an optimization method.
We updated the parameters for 800,000 iterations on both datasets. We experimented with different numbers of patch divisions $N$, specifically $N=2,4,8$, and also trained the conventional DPM without patch partitioning as the original method.

\vspace{1mm}\noindent{}\textbf{Evaluation Metrics}~
We used Fréchet Inception Distance (FID)~\cite{FID, clean-fid}, a metric that measures the distance between two image distributions and provides a measure of generative quality and diversity.
For CelebA, we used 40K generated images and 40K real images to measure the FID score.
For LSUN bedroom, we used 50K generated images and 50K real images.
Real images were selected randomly from each test set.
Additionally, we used the maximum memory consumption during inference as another evaluation metric.

\subsection{Qualitative Evaluation}
The results of the proposed method trained on CelebA dataset are presented in \cref{fig:sampled_CelebA}.
When $N=2$ or $4$, the generated images appear natural.
However, when $N=8$, the boundary lines between patches become prominent, even though content information is shared among the patches.
Similarly, \cref{fig:sampled_LSUN} shows the results of the proposed method on LSUN bedroom dataset.
The generated images look natural when $N=2$, but when $N\geq4$, patch boundaries become noticeable.
While the network is conditioned with location information, the conditioning is not precise enough to achieve seamless patch connection at the boundaries, where accurate location information is essential.

\subsection{Quantitative Evaluation}
\begin{table}[t]
    \centering
    \caption{FID~($\downarrow$) evaluation of our proposed methods and the original. The lower is better.}
    \vspace{-3mm}
    \begin{tabular}{ccc}
        \bhline{1.0pt}  
        Dataset                        & Division $N$ & FID~($\downarrow$) \\ \hline
        \multirow{4}{*}{CelebA}       & 1 (original)  & 14.8                        \\
                                      & 2         & 14.0                        \\
                                      & 4         & 13.6                        \\
                                      & 8         & 16.2                        \\ \hline
        \multirow{4}{*}{LSUN bedroom} & 1 (original)  & 22.4                         \\
                                      & 2         & 24.1                        \\
                                      & 4         & 28.8                        \\
                                      & 8         & 66.1                         \\ \bhline{1.0pt}
    \end{tabular}
    \label{table:FID}
    \vspace{-1mm}
\end{table}
\begin{table}[t]
    \centering
    \caption{Maximum memory consumption of the original method and our proposed method at different number of divisions $N$.}
    \vspace{-3mm}
    \begin{tabular}{cc}
        \bhline{1.0pt}  
        Division $N$ & Max memory consumption {[GB]} \\ \hline
        1 (original)   & 7.46               \\
        2          & 3.25               \\
        4          & 2.14               \\
        8          & 1.83      \\  \bhline{1.0pt}
    \end{tabular}
    \label{table:memory_consumption}
    \vspace{-3mm}
\end{table}
The proposed method's FID results are presented in \cref{table:FID} for both CelebA and LSUN bedroom datasets.
On the CelebA dataset, when $N=2$ or $4$, the proposed method's performance is slightly better than the original method.
However, as the number of patch divisions $N$ increases up to eight, the FID score inclines sharply, consistent with the qualitative evaluation results.
Similarly, for the LSUN bedroom dataset, the performance of the proposed method is not significantly different from that of the original method when $N=2$.
However, the FID deteriorates sharply when the number of divisions increases over four or more.

The maximum memory consumption of the original and proposed methods is presented in \cref{table:memory_consumption}.
As the computation part of the reverse process is the same for both methods, only the forward propagation part of the network was considered for measuring the maximum memory consumption.
The proposed method with $N=2$ reduced the memory consumption by over 50\% compared to the original method.
When $N\geq4$, the memory consumption decreases as $N$ increases, although not uniformly due to the increased number of input channels caused by GCC.
\cref{fig:FID_vs_memory} shows the trade-off relation between maximum memory consumption and the FID score.
Overall, the proposed method effectively reduces the maximum memory consumption during inference, preserving comparable generative quality.

    \section{Discussion}
    \label{sec:discussion}
For LSUN betroom dataset, significant degradation in quality was observed when the number of divisions $N$ became four or more. 
In addition, comparing the results of CelebA and LSUN bedroom datasets for $N=8$ divisions, LSUN bedroom showed a larger quality degradation compared to the original method. 
This is because the LSUN bedroom dataset includes various compositions and may be difficult to train compared to the CelebA dataset, which has a uniform composition. 
To overcome this issue, data augmentation techniques such as affine transformation proposed by Karras et al.~\cite{EDM} may be useful for datasets with non-uniform compositions. 

Currently, GCC extracts content information from the entire input image at every diffusion step, which may lead to error accumulation during the reverse process and contribute to discontinuities on the boundaries between adjacent patches.
To address this issue, further improvements can be made, such as limiting the extraction of the entire content information to only one time.

    \section{Conclusion}
    \label{sec:conclusion}
    We introduced a patch-based DPM, a memory-efficient diffusion probabilistic model generating images on a patch-by-patch basis through positional embedding and Global Content Conditioning.
The experiments showed that our approach could reduce the maximum memory consumption by half while maintaining comparable image quality when the entire image was divided into $2\times2$ patches.
    
    \noindent\\{\bf Acknowledgements}
    \label{sec:acknowledgements}
    {\footnotesize
        We thank Asst. Prof. Qi Feng for valuable feedbacks as well as English proofreading and Yoshiki Kubotani for feedbacks regarding the graphical design.
This research is supported by the JSPS KAKENHI Grant Number 21H05054.
    }
    
    {\small
        \bibliographystyle{ieee_fullname}
        \bibliography{bibliography}

\begin{thebibliography}{10}\itemsep=-1pt

\bibitem{ILVR}
Jooyoung Choi, Sungwon Kim, Yonghyun Jeong, Youngjune Gwon, and Sungroh Yoon.
\newblock {ILVR}: Conditioning method for denoising diffusion probabilistic
  models.
\newblock In {\em ICCV}, pages 14347--14356, 2021.

\bibitem{ADM}
Prafulla Dhariwal and Alexander Nichol.
\newblock Diffusion models beat {GAN}s on image synthesis.
\newblock In {\em NeurIPS}, volume~34, pages 8780--8794, 2021.

\bibitem{GAN}
Ian Goodfellow, Jean Pouget-Abadie, Mehdi Mirza, Bing Xu, David Warde-Farley,
  Sherjil Ozair, Aaron Courville, and Yoshua Bengio.
\newblock Generative adversarial nets.
\newblock In {\em NeurIPS}, volume~27, 2014.

\bibitem{FID}
Martin Heusel, Hubert Ramsauer, Thomas Unterthiner, Bernhard Nessler, and Sepp
  Hochreiter.
\newblock {GAN}s trained by a two time-scale update rule converge to a local
  nash equilibrium.
\newblock In {\em NeurIPS}, volume~30, 2017.

\bibitem{DDPM}
Jonathan Ho, Ajay Jain, and Pieter Abbeel.
\newblock Denoising diffusion probabilistic models.
\newblock In {\em NeurIPS}, volume~33, pages 6840--6851, 2020.

\bibitem{CelebA-HQ}
Tero Karras, Timo Aila, Samuli Laine, and Jaakko Lehtinen.
\newblock Progressive growing of {GAN}s for improved quality, stability, and
  variation.
\newblock In {\em ICLR}, 2018.

\bibitem{EDM}
Tero Karras, Miika Aittala, Timo Aila, and Samuli Laine.
\newblock Elucidating the design space of diffusion-based generative models.
\newblock In {\em NeurIPS}, 2022.

\bibitem{StyleGAN}
Tero Karras, Samuli Laine, and Timo Aila.
\newblock A style-based generator architecture for generative adversarial
  networks.
\newblock In {\em CVPR}, June 2019.

\bibitem{Adam}
Diederik~P. Kingma and Jimmy Ba.
\newblock Adam: {A} method for stochastic optimization.
\newblock In {\em ICLR}, 2015.

\bibitem{VAE}
Diederik~P. Kingma and Max Welling.
\newblock {Auto-Encoding Variational Bayes}.
\newblock In {\em ICLR}, 2014.

\bibitem{CIFAR10}
Alex Krizhevsky.
\newblock Learning multiple layers of features from tiny images.
\newblock Technical report, 2009.

\bibitem{COCO-GAN}
Chieh~Hubert Lin, Chia-Che Chang, Yu-Sheng Chen, Da-Cheng Juan, Wei Wei, and
  Hwann-Tzong Chen.
\newblock {COCO-GAN}: Generation by parts via conditional coordinating.
\newblock In {\em ICCV}, pages 4511--4520, 2019.

\bibitem{CelebA}
Ziwei Liu, Ping Luo, Xiaogang Wang, and Xiaoou Tang.
\newblock Deep learning face attributes in the wild.
\newblock In {\em ICCV}, pages 3730--3738, 2015.

\bibitem{SDEdit}
Chenlin Meng, Yutong He, Yang Song, Jiaming Song, Jiajun Wu, Jun-Yan Zhu, and
  Stefano Ermon.
\newblock {SDE}dit: Guided image synthesis and editing with stochastic
  differential equations.
\newblock In {\em ICLR}, 2022.

\bibitem{ImprovedDDPM}
Alexander~Quinn Nichol and Prafulla Dhariwal.
\newblock Improved denoising diffusion probabilistic models.
\newblock In {\em Proceedings of the 38th International Conference on Machine
  Learning}, volume 139, pages 8162--8171, 2021.

\bibitem{clean-fid}
Gaurav Parmar, Richard Zhang, and Jun-Yan Zhu.
\newblock On aliased resizing and surprising subtleties in gan evaluation.
\newblock In {\em CVPR}, 2022.

\bibitem{StyleCLIP}
Or Patashnik, Zongze Wu, Eli Shechtman, Daniel Cohen-Or, and Dani Lischinski.
\newblock Style{CLIP}: Text-driven manipulation of stylegan imagery.
\newblock In {\em ICCV}, pages 2085--2094, October 2021.

\bibitem{LDM}
Robin Rombach, Andreas Blattmann, Dominik Lorenz, Patrick Esser, and Björn
  Ommer.
\newblock High-resolution image synthesis with latent diffusion models.
\newblock In {\em CVPR}, pages 10674--10685, 2022.

\bibitem{unet}
Olaf Ronneberger, Philipp Fischer, and Thomas Brox.
\newblock {U-Net}: Convolutional networks for biomedical image segmentation.
\newblock In {\em Proceedings of Medical Image Computing and Computer-Assisted
  Intervention}, pages 234--241, 2015.

\bibitem{SR3}
Chitwan Saharia, Jonathan Ho, William Chan, Tim Salimans, David~J. Fleet, and
  Mohammad Norouzi.
\newblock Image super-resolution via iterative refinement.
\newblock {\em IEEE Transactions on Pattern Analysis and Machine Intelligence},
  45(4):4713--4726, 2023.

\bibitem{LION5B}
Christoph Schuhmann, Romain Beaumont, Richard Vencu, Cade~W Gordon, Ross
  Wightman, Mehdi Cherti, Theo Coombes, Aarush Katta, Clayton Mullis, Mitchell
  Wortsman, Patrick Schramowski, Srivatsa~R Kundurthy, Katherine Crowson,
  Ludwig Schmidt, Robert Kaczmarczyk, and Jenia Jitsev.
\newblock {LAION}-5{B}: An open large-scale dataset for training next
  generation image-text models.
\newblock In {\em 36th Conference on Neural Information Processing Systems
  Datasets and Benchmarks Track}, 2022.

\bibitem{original_DDPM}
Jascha Sohl-Dickstein, Eric~A. Weiss, Niru Maheswaranathan, and Surya Ganguli.
\newblock Deep unsupervised learning using nonequilibrium thermodynamics.
\newblock In {\em Proceedings of the 32nd International Conference on Machine
  Learning}, volume~37, pages 2256--2265, 2015.

\bibitem{DDPM++}
Yang Song, Jascha Sohl-Dickstein, Diederik~P Kingma, Abhishek Kumar, Stefano
  Ermon, and Ben Poole.
\newblock Score-based generative modeling through stochastic differential
  equations.
\newblock In {\em ICLR}, 2021.

\bibitem{transformer}
Ashish Vaswani, Noam Shazeer, Niki Parmar, Jakob Uszkoreit, Llion Jones,
  Aidan~N Gomez, \L~ukasz Kaiser, and Illia Polosukhin.
\newblock Attention is all you need.
\newblock In {\em NeurIPS}, volume~30, 2017.

\bibitem{LSUN}
Fisher Yu, Yinda Zhang, Shuran Song, Ari Seff, and Jianxiong Xiao.
\newblock {LSUN:} construction of a large-scale image dataset using deep
  learning with humans in the loop.
\newblock {\em CoRR}, abs/1506.03365, 2015.

\end{thebibliography}
    }
\end{document}